\documentclass[runningheads]{llncs}
\usepackage[T1]{fontenc}
\usepackage{graphicx}
 
\usepackage{caption}
\usepackage{subcaption}
\usepackage{amsmath}
\usepackage{amsfonts}
\usepackage{amssymb}
\usepackage{bbm}
\usepackage{bm}
\usepackage{lipsum}
\usepackage{stackengine}
\usepackage{tabularx}
\usepackage{makecell}
\usepackage{booktabs}
\usepackage{wrapfig}

\usepackage{tikz}
\usetikzlibrary{decorations.pathreplacing,calligraphy,calc} %
\usepackage{xcolor}
\definecolor{darkgreen}{RGB}{9, 133, 9}
\usepackage{pgfplots}
\pgfplotsset{compat = 1.9}
\usepgfplotslibrary{external}
\usepackage{mathtools}
\mathtoolsset{showonlyrefs=true}

\usepackage{algorithm}
\usepackage{algpseudocode}
\usepackage{optidef}

\usepackage{multirow}

\usepackage[colorinlistoftodos,prependcaption,textsize=tiny,textwidth=20mm]{todonotes}

\newcommand\fra[1]{\textcolor{black}{#1}}

\DeclareMathOperator{\Tr}{Tr}

\usepackage{xspace}
\newcommand{\nlevels}{n_\text{levels}\xspace}
\newcommand{\matr}[1]{{#1}}     %

\begin{document}
\title{Combining Primal and Dual Representations in Deep Restricted Kernel Machines Classifiers}
\titlerunning{Combining Primal and Dual in DRKM Classifiers}
\author{Francesco Tonin \and
Panagiotis Patrinos \and
Johan A. K. Suykens}
\institute{ESAT-STADIUS, KU Leuven, Heverlee 3001, Belgium.\\
\email{\{francesco.tonin,panos.patrinos,johan.suykens\}@esat.kuleuven.be}}
\maketitle              %

\begin{abstract}
\fra{In the context of deep learning with kernel machines}, the deep Restricted Kernel Machine (DRKM) framework allows multiple levels of kernel PCA (KPCA) and Least-Squares Support Vector Machines (LSSVM) to be combined into a deep architecture using visible and hidden units. 
We propose a new method for DRKM classification coupling the objectives of KPCA and classification levels, with the hidden feature matrix lying on the Stiefel manifold.
The classification level can be formulated as an LSSVM or as an MLP feature map, combining depth in terms of levels and layers. 
The classification level is expressed in its primal formulation, as the deep KPCA levels\fra{, in their dual formulation,} can embed the most informative components of the data in a much lower dimensional space.
\fra{The dual setting is independent of the dimension of the inputs and the primal setting is parametric, which makes the proposed method computationally efficient for both high-dimensional inputs and large datasets.}
In the experiments, \fra{we show that our developed algorithm can effectively learn from small datasets}, while using less memory than the convolutional neural network (CNN) with high-dimensional data.
and that models with multiple KPCA levels can outperform models with a single level.
\fra{On the tested larger-scale datasets, DRKM is more energy efficient than CNN while maintaining comparable performance.}
\end{abstract}

\section{Introduction}
The deep Restricted Kernel Machine (DRKM) was introduced in \cite{drkm} to find synergies between kernel methods and deep learning. While recent research has focused on leveraging RKM in generative models \cite{joachim2018,pandey2021} and unsupervised learning \cite{tonin2020,tonin2021,tonin2023}, there has been little investigation into the use of DRKM for classification problems. 
The \textit{single}-level RKM classification framework was used in \cite{houthuys2021} to propose a multi-modal classification method based on tensor learning using a single model weight tensor with all the modes sharing a common latent space. 
The \textit{semi}-supervised RKM was used in conjunction with message-passing kernel PCA (KPCA) levels in \cite{achten2023} for semi-supervised node classification in graphs using \textit{only} the dual representation.
In \cite{drkm}, deep reduced set kernel-based models were applied to classification datasets, and an unconstrained DRKM with a primal-estimation scheme using a Least-Squares Support Vector Machines (LSSVM) classifier was evaluated on two UCI datasets. However, its unconstrained formulation introduces additional stabilization terms in the objective and does not combine primal and dual in the same model.
A DRKM for multi-level kernel PCA (DKPCA) was introduced in \cite{tonin2023} as a method for unsupervised representation learning employing multiple levels of kernel PCA with orthogonality constraints. However, this method does not consider a classification objective and only employs the dual representation.
Overall, effective classification algorithms for deep (i.e., with multiple levels) RKMs are still an open problem and previous studies have only considered either primal or dual models.%

In this paper, we propose a new DRKM classifier employing multiple levels of KPCA and an LSSVM or MLP classification level.
We combine the dual representation of the KPCA with the primal representation of the LSSVM/MLP: the dual representation is better suited for the kernel trick and feature extraction, while the primal representation is more computationally efficient for larger datasets and can directly take advantage of the nonlinear embedding from the multiple KPCA levels.
\fra{Given that the  size of the  kernel matrix is independent of the dimension of the inputs, the dual problem is typically suited for handling problems with high-dimensional input data, while solving the primal is often suitable for large numbers of data. Combining the parametric primal representation  and the kernel-based  dual in the DRKM provides a  powerful  framework  for  efficient algorithms for both high-dimensional inputs and larger-scale datasets.
Therefore, the model is well-suited for Frugal AI, where the non-parametric levels can handle high-dimensional data with computational resources that are independent of the number of inputs, possibly reducing the dimensionality of the input space, such that the parametric classification level can be efficiently trained in a much lower dimensional space.
Note that we employ the dual framework for the unsupervised KPCA levels and the primal framework for the supervised classification level. Employing unsupervised levels is advantageous when training with limited labelled data, as the model is built on unsupervised core models that can effectively induce patterns and representations from the unlabelled data.}
Our main contributions are as follows.
\begin{itemize}
    \item We propose a new method for DRKM classification that couples the objectives of KPCA and LSSVM/MLP classification levels to create a deep architecture. Our method integrates the kernel-based dual representation in the unsupervised part and the parametric primal representation in the supervised part. \fra{In this way, the final model is applicable to general data in the sense that the DRKM can be applied to limited and larger amounts of data in either low or high dimensional inputs spaces.}
    \item We define a constrained optimization problem with orthogonality constraints on the hidden features of the unsupervised levels and illustrate a training algorithm based on Projected Gradient Descent. In particular, we investigate how end-to-end training compares with unsupervised initialization of the KPCA levels followed by fine-tuning.
    \item We empirically evaluate our method on benchmark datasets comparing with the LSSVM/MLP classifiers alone, showing that increased depth can benefit performance in kernel methods when limited training data is available. \fra{Our approach is frugal in that it can effectively learn from limited training data thanks to its unsupervised KPCA core, and it can also efficiently deal with larger datasets thanks to the primal classification level. We additionally empirically show lower environmental impact than the CNN on a number of UCI datasets with thousands of dimensions, thanks to the employed non-parametric dual representation, with similar classification performance.}
\end{itemize}

\section{Background}
Consider the training data $\{(x_i, y_i)\}^N_{i=1}$, where $x_i \in \mathbb{R}^d$ and $y_i \in \{-1, 1\}$. The objective of RKM binary classification \cite{drkm} is:
\begin{equation} \label{eq:original}
	J_0 = \sum_{i=1}^N \left( 1-y_i\left(\varphi(x_i)^Tw+b\right)\right)h_i - \frac{\lambda}{2} h_i^2 + \frac{\eta}{2} w^Tw,
\end{equation}
where $\varphi(x) \in \mathbb{R}^l$ is the feature map, $w \in \mathbb{R}^l$ is the weight vector, $h_i \in \mathbb{R}$ are the hidden features, and $\lambda, \eta > 0$ are regularization constants. 
\fra{It can be shown that \eqref{eq:original} is a lower bound to the LSSVM classification objective \cite{suykens2002,drkm}.}
In the shallow case where there is no other level, the solution of \eqref{eq:original} in the conjugated features $h_i$ is given by a linear system obtained from the stationarity of $J_0$ \cite{drkm}. \\

Regarding the KPCA level, \fra{let $s$ be the number of selected principal components. In the LS-SVM setting, the KPCA problem can be written as minimizing a regularization term and finding directions of maximum variance \cite{suykens2002}:
\begin{mini}|l|
	{\matr{W},e_i}{\tilde{J}_{\text{KPCA}} = \frac{\eta}{2}\Tr{(\matr{W}^\top \matr{W})}-\frac{1}{2}\sum_{i=1}^N e_i^\top \Lambda^{-1}  e_i}{}{}
	\label{eq:kpca}
	\addConstraint{ e_i}{= \matr{W}^\top \varphi( x_i),}{\quad i=1,\dots,N,}
\end{mini}
where $\matr{W} \in \mathbb{R}^{l \times s}$ is the interconnection matrix, $ e_i \in \mathbb{R}^{s}$ are the score variables along the selected $s$ projection directions, and $\matr{\Lambda}={\rm diag}\{\lambda_1, \ldots, \lambda_s\} \succ 0, \eta > 0$ are regularization hyperparameters. 
The RKM formulation of KPCA \cite{drkm} is given by an upper bound of $\tilde{J}_{\text{KPCA}}$ obtained component-wise with the Fenchel-Young inequality 
$\frac{1}{2\lambda}e^2+\frac{\lambda}{2}h^2 \geq eh, \, \forall e, h \in \mathbb{R}$
which introduces the hidden features $ h$}  and leads to the following objective with \textit{conjugate feature duality}:
\begin{equation} \label{eq:rkm-kpca}
	{J}_{\rm KPCA} = -\sum_{i=1}^N \varphi(  x_i)^\top  \matr{W}  h_i+\frac{1}{2}\sum_{i=1}^N { h_i}^\top\matr{\Lambda}  \ h_i+\frac{\eta}{2}\Tr{\left( \matr{W}^\top \matr{W} \right)},
\end{equation}
where $ h_i\in \mathbb R^s$ are the conjugated hidden features corresponding to each training sample $x_i$; in representation learning, $ h_i$ is also known as the latent representation of $ x_i$ consisting of $s$ latent variables or of $s$ hidden features.
\fra{By characterizing the stationary points of ${J}_{\rm KPCA}$ in \eqref{eq:rkm-kpca}, the following eigenvalue problem is obtained
\begin{equation}\label{eq:eigen:kpca}
	\frac{1}{\eta} \matr{K} \matr{H} = \matr{H} \matr{\Lambda},
\end{equation}
where $\matr{K}\in \mathbb R^{N\times N}$ denotes the kernel matrix induced by the positive-definite kernel function $k: \mathbb{R}^d \times \mathbb{R}^d \mapsto \mathbb{R}$ with $k( x_i, x_j)=\varphi( x_i)^\top \varphi( x_j)$  and  the matrix $\matr{H} = [ h_1, \dots,  h_N]^\top$ incorporates the conjugate hidden features for all $N$ data points.}
Since the primal form in \eqref{eq:rkm-kpca} is not directly suitable for minimization because it is unbounded below, \cite{tonin2020} derived a rewriting of \eqref{eq:rkm-kpca} in the dual by introducing orthogonality constraints on the hidden features and eliminating the interconnection matrix. Using the stationarity of $J_{\text{KPCA}}$ and the kernel trick, the first term of \eqref{eq:rkm-kpca} can be rewritten as \fra{$-\frac{1}{\eta} \Tr{\left( H^T K H \right)}$, and the third term can be rewritten as $\frac{1}{2\eta}\Tr{\left( H^T K H \right)}$. Derivation details are given in \cite{tonin2020}}. 
A deep KPCA (DKPCA) framework combining multiple KPCA levels in a single objective was proposed in \cite{tonin2023}. In that work, the principal components of multiple KPCA levels are coupled in their hidden units creating both forward and backward dependency across levels, showing greater representation efficiency for unsupervised learning. In this work, we propose a new method for classification based on a DKPCA architecture coupled with a classification level with end-to-end training combining dual and primal representations.

\subsection{Related Work}
\fra{
Deep kernel learning tackles multiple latent spaces for greater flexibility, more informative hierarchical investigation of the data, and kernel-based interpretations. 
\cite{montavon2011kernel} considers the representation learned by successive network layers.
In particular, they consider a deep neural network (DNN) of $L$ layers $f(x)=f_L \circ \dots \circ f_1(x)$ and analyze the induced kernels $k_0=k_\text{RBF}(x,x'), k_1=k_\text{RBF}(f_1(x),f_1(x')), \dots$, $k_L=k_\text{RBF}(f_L \circ \dots \circ f_1(x),f_L \circ \dots \circ f_1(x'))$, with $k_\text{RBF}$ indicating the Gaussian kernel. In other words, each layer of the DNN is associated with a kernel defined on the output of that layer. 
Conversely, in this paper we consider several feature maps over multiple levels. 
In other words, following the terminology used in \cite{drkm}, in \cite{montavon2011kernel} deep learning is only performed over \textit{layers}, while in our approach depth is given by multiple \textit{levels}, each associated with a different feature map possibly consisting of multiple layers.
For instance, a DRKM can consist of multiple DNNs, each associated with the feature map $\varphi_j$ of level $j$. Therefore, in the framework of DRKM, \cite{montavon2011kernel} performs shallow learning, as it works with a single DNN.
Additionally, in the DRKM the levels are coupled in terms of their hidden features, while in \cite{montavon2011kernel} the considered architecture is a composition of the outputs of each layer.
A concatenation of operator-valued kernel layers was considered for data autoencoding in \cite{laforgue2019}, but no extension to supervised learning was considered. 
In \cite{deng2019}, shallow PCA is conducted to extract principal components, which are then applied to another KPCA, where each KPCA independently and sequentially optimizes its variance maximization. PCA is firstly performed to extract principal components of the data and then further dimensionality reduction is sequentially applied to the extracted features from the previous (K)PCA layer. This serial approach makes each layer straightforwardly maximize its variance objective, which is independent of other layers.
}

\section{Proposed Method}
In this section, we describe the proposed combination of KPCA levels in the dual and a classification level in the primal. The former part extracts multiple levels of the most informative components of the given data through multiple feature maps. After such transformations, the classification level can effectively learn a decision boundary in a much lower dimensional subspace, motivating its representation in the primal. We start by describing the model formulation where the primal classifier is an LSSVM classifier following one ore more KPCA levels. Next, we derive the model for the MLP classifier. Finally, we discuss the optimization algorithm and  multiple initialization procedures.

\subsection{Shallow RKM - Primal LSSVM Classifier}

We start by defining a DRKM classifier with a single KPCA level, i.e., the shallow RKM. Its architecture is the following.
\begin{itemize}
\item Level 1 consists of KPCA using as input the sample $x_i$. The features extracted by this level are characterized by its hidden features $h_i^{(1)}$.
\item Level 2 consists of LSSVM classification using as input the hidden features $h_i^{(1)}$ from the previous level and with output data $y_i$. This level is characterized by its weights $w$.
\end{itemize}
The training objective of the above architecture is obtained from the representation learning optimization problem of RKM with one KPCA level and an LSSVM classifier written in its RKM formulation. The solution in terms of the $h_i$ of the latter is given by a linear system in the shallow case; however, in the case of multiple levels, end-to-end learning cannot be performed by solving the linear system. We propose an alternative training strategy for the deep case. First of all, we eliminate $h_i$ in \eqref{eq:original}: from the stationarity of $J_0$ it follows that
\begin{equation} \label{eq:h_i}
    h_i = \frac{1}{\lambda} \left( 1-y_i\left(w^T\varphi(x_i)+b\right) \right).
\end{equation}

Replacing \eqref{eq:h_i} in \eqref{eq:original},
\begin{equation} \label{eq:original_wo_h}
    J_0 = \sum_{i=1}^N \frac{1}{2\lambda} \left( 1-y_i\left(w^T\varphi(x_i)+b\right) \right)^2 + \frac{\eta}{2} w^Tw.
\end{equation}

We propose to combine one level of KPCA in the RKM formulation with the objective of an LSSVM classifier in the following optimization problem:
\begin{equation} \label{eq:onelevel}
\begin{alignedat}{3}
&\stackunder{min}{$h_i^{(1)},w,b$}  &\,&  J_1=-\frac{1}{2\eta_1} \Tr \left({H^{(1)}}^T K^{(0)} H^{(1)} \right) + \frac{\eta}{2} w^Tw &\\
& & & + \frac{1}{2\lambda} \sum_{i=1}^N \left(1-y_i(w^T\varphi(h_i^{(1)})+b)\right)^2      &\\
&\quad \, \text{s.t.} &      & {H^{(1)}}^T H^{(1)} = I.                     &
\end{alignedat}
\end{equation}
The objective of \eqref{eq:onelevel} consists of a KPCA term written in the dual formulation and a classification term in the primal. The former problem is independent of the input dimensionality due to the dual formulation; we use the primal formulation for the latter term because $h_i^{(1)}$ is usually low-dimensional, as the number of principal components $s_1$ is generally small. The classifier is expressed in the primal as
$
	\hat{y} = \text{sign} \left( w^T \varphi(h^{(1)}) + b \right),
$
where $\hat{y}$ is the estimated label of an input sample $x$ and $h_i$ is computed following \eqref{eq:h_i} with input $h^{(1)}$, which is the latent representation of $x$. Regarding the choice of $\varphi$, one can employ a linear mapping as often done in the last layer of deep neural network classifiers; the full model is still non-linear due to the first KPCA level. Moreover, the formulation in \eqref{eq:onelevel} is intended for binary classification problems; multiclass problems with $p$ classes can be solved with a one-vs-all approach by training $p$ binary classifiers.

\begin{remark}[Orthogonality constraints]
Contrary to the classifier proposed in \cite{drkm}, we use orthogonality constraints on the hidden features of the kernel PCA level. In the field of neural networks, it was shown that orthogonality constraints can lead to better performance \cite{bansal2018} and better regularization \cite{cogswell2016,cho2017}. For example, \cite{cho2017} shows that, if the weights of the neural network are points of the Grassmann manifold $\mathcal{G}(1,n)$, minimizing the soft orthogonality constraint $ L(\alpha, W) = \frac{\alpha}{2}|| W^TW-I ||^2_F,$ where $W \in \mathbb{R}^{n \times p}$ is the weight matrix of the neural network, also minimizes a model complexity loss, introduced in \cite{graves2011}, based on the KL divergence $D_\text{KL} (Q(w|\beta) || P(w|\alpha))$ between the posterior $Q(w|\beta)$ and the prior $P(w|\alpha)$ distribution of the neural network weights. In contrast to \cite{cho2017}, our method is based on kernel methods.
\end{remark}

\subsection{Deep RKM - Primal LSSVM Classifier}

\begin{figure}[t]
\centering
\includegraphics{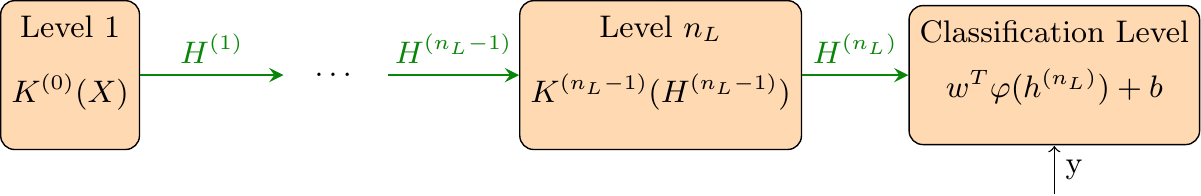}
\caption{\fra{Diagram of the proposed DRKM of \eqref{eq:nlevels} with $n_L$ KPCA levels in dual and one final LSSVM classification level in primal. Each green arrow goes from the level that is characterized by the corresponding hidden features to the level where it is used as input. $\matr{K}^{(0)}(\matr{X})$ indicates the kernel matrix of the input data.}}
\label{fig:dual}
\end{figure}

We now define the DRKM classifier with $\nlevels$ KPCA levels: increased classifier depth may lead to a more compact representation of the target function and thus to better performance for complex classification problems. The architecture, \fra{illustrated in Fig. \ref{fig:dual}}, is as follows.
\begin{itemize}
\item Level 1 consists of KPCA using as input the sample $x_i$. The features extracted by this level are characterized by its hidden features $h_i^{(1)}$.
\item Level $j, \, 2 \leq j \leq \nlevels$ consists of KPCA using as input the hidden features $h_i^{(j-1)}$ from the previous level. The features extracted by this level are characterized by its hidden features $h_i^{(j)}$.
\item Level $\nlevels+1$ consists of LSSVM classification using as input the hidden features $h_i^{(\nlevels)}$ from the previous level and with output data $y_i$. This level is characterized by its weights $w$.
\end{itemize}

The proposed objective to be minimized for $n_\text{levels} \geq 2$ of KPCA levels is:

\begin{align} \label{eq:nlevels}
& J_{n_\text{levels}}(h_i^{(1)},\dots,h_i^{(n_\text{levels})},w,b) =
 -\frac{1}{2\eta_1} \Tr \left({H^{(1)}}^T K^{(0)} H^{(1)} \right) \\
& \qquad - \sum_{j=2}^{n_\text{levels}} \frac{1}{2\eta_j} \Tr \left({H^{(j)}}^T K^{(j-1)} (H^{(j-1)}) H^{(j)} \right) \\
& \qquad + \frac{1}{2\lambda} \sum_{i=1}^N \left(1-y_i(w^T  \varphi(h_i^{(n_\text{levels})})+b)\right)^2 + \frac{\eta}{2} w^Tw,
\end{align}
where \fra{$\matr{H}^{(j)} = [h_1^{(j)}, \dots, h_N^{(j)}]^\top \in \mathbb R^{N \times s_j}$ incorporates the hidden features conjugated along $s_j$ projection directions for all $N$ data points, where $s_j$ is the number of selected principal components by the $j$-th level}.
\fra{The kernel matrices are obtained as follows: $\matr{K}^{(0)} \in \mathbb{R}^{N \times N}$ is attained as $(K^{(0})_{ik} = k^{(0)}(x_i, x_k)$ and $\matr{K}^{(j-1)} \in \mathbb{R}^{N \times N}$ as  $(K^{(j-1)})_{ik} = k^{(j-1)}(  h_i^{(j-1)}, h_k^{(j-1)})$, where  $k^{(0)}: \mathbb{R}^d \times \mathbb{R}^d \mapsto \mathbb{R}$ is the kernel function of the first level and $k^{(j-1)}: \mathbb{R}^{s_{j-1}} \times \mathbb{R}^{s_{j-1}} \mapsto \mathbb{R}$ is the kernel function of level $j=2,\dots,{n_\text{levels}}$.}
\fra{The projection directions of shallow KPCA are uncorrelated due to the orthogonality of different principal components; similarly, we impose ${H^{(j)}}^\top H^{(j)}=I$ in \eqref{eq:nlevels}.}
\fra{Following \cite{tonin2020}, orthogonality between the hidden features of two different levels can be imposed to encourage the levels to learn new features of the data instead of repeating the same features in every level.}
\fra{The number of levels is an hyperparameter to be tuned in addition to the kernel hyperparameters and the number of components in each level. One can tune these hyperparameters using standard techniques such as defining a separate validation set or employing cross-validation. In general, the minimum number of levels required for good performance on a given set of observations depends on the complexity of the training dataset.}
For an out-of-sample $x^\star$, multiple strategies to compute its latent representation have been proposed in the context of RKMs \cite{drkm,arun2,tonin2020,tonin2023}. Based on \cite{joachim2018}, we propose to apply a kernel smoother approach \cite{hastie2009}
$
    {h^{(\nlevels)}}^\star = \dfrac{\sum_{i=1}^N \tilde{k}(x_i, x^\star) h_i^{(\nlevels)}}{\sum_{i=1}^N \tilde{k}(x_i, x^\star)},
$
where $\tilde{k}$ is a localized kernel such as the Gaussian kernel with design parameter $\tilde{\sigma}$, representing the width of the considered local neighborhood. 
A larger $\tilde{\sigma}$ gives lower variance, as more observed points are considered, but also higher bias. With smaller bandwidth, the similarity measure expressed by $\tilde{k}$ is more local, so ${h^{(\nlevels)}}^\star$ is closer to the latent representation of training points that are similar to $x^\star$. Using this approach, the predicted class of $x^\star$ expressed in the primal model representation is:
\begin{equation} \label{eq:classifier-kernelsmoother}
    \hat{y}^\star = \text{sign} \left[ w^T \varphi \left( %
		\frac{\sum_{i=1}^N \tilde{k}(x_i, x^\star) h_i^{(\nlevels)}}{\sum_{i=1}^N \tilde{k}(x_i, x^\star)} \right)
    + b \right].
\end{equation}
Note that $h_i^{(\nlevels)}$ depends on the hidden features of the previous levels $h_i^{(\nlevels-1)}, \allowbreak \dots, \allowbreak h_i^{(1)}$ as they are jointly optimized in \eqref{eq:nlevels}. 
\fra{The final architecture of the proposed DRKM is therefore a combination of non-parametric modelling through the hidden features of the last KPCA level computed through \eqref{eq:classifier-kernelsmoother} and the parametric classifier with parameters $w, b$.}

\subsection{Deep RKM - Primal MLP Classifier}
The classification DRKM can also be constructed with multiple KPCA levels in the RKM formulation and a final MLP classification level, instead of the LSSVM classifier described in the previous subsection. 
The MLP classifier is a single level that consists of a multi-layer feature map.
Note here the difference between depth given by multiple KPCA levels and depth in the MLP: in the latter case deep learning is only performed over layers, while in the former case depth is given by multiple levels, each associated with a different feature map possibly consisting of multiple layers.

The training objective of the DRKM with a final MLP $f_\theta(h_i^{(\nlevels)}): \mathbb{R}^{s_{\nlevels}} \to \mathbb{R}^p$ classification level with $p$ classes is:

\begin{align} \label{eq:mlp}
& J(h_i^{(1)},\dots,h_i^{(n_\text{levels})},W,b,\theta) = -\frac{1}{2\eta_1} \Tr \left({H^{(1)}}^T K^{(0)} H^{(1)} \right)\\
& \qquad - \sum_{j=2}^{n_\text{levels}} \frac{1}{2\eta_j} \Tr \left({H^{(j)}}^T K^{(j-1)} (H^{(j-1)}) H^{(j)} \right) \\
& \qquad + \frac{1}{2\lambda N} \sum_{i=1}^N \mathcal{L}(f_\theta(h_i^{(n_\text{levels})}), y_i) + \frac{\eta}{2} \Tr W^TW,
\end{align}
where 
$\mathcal{L}(\hat{y}_i, y_i) = -\sum_{k=1}^p y_{i,k} \log \frac{\exp{\hat{y}_{i,k}}}{\sum_{j=1}^p \exp{\hat{y}_{i,j}}}$ is the cross entropy loss with $\hat{y}_i = f_\theta(h_i^{(n_\text{levels})})$ and $y_{i,k}$ a binary indicator of input $x_i$ belonging to class $k$.

The logits predicting the class of an out-of-sample point $x^\star$ are given by $f_\theta(h^{(\nlevels)})$, where $h^{(\nlevels)}$ is obtained through the kernel smoother approach. 
\fra{One motivation to employ the MLP classification level is that} the MLP approach also naturally handles the multi-class case, resulting in multiple logits, one for each class.
The next subsection describes the optimization algorithm for the DRKM classifier consisting of multiple KPCA levels and a final LSSVM or MLP classification level.

\subsection{Optimization}

The nonlinear optimization problem with objective \eqref{eq:nlevels} or \eqref{eq:mlp} has at least one global minimum due to the Weierstrass theorem, as the objective function is continuous and the feasible set is compact, and it may have multiple local minima since it is a non-convex problem. The constraint set is a Stiefel manifold $\text{St}(s,N)$, so one of the algorithms that have been proposed for optimization on the Stiefel manifold could be employed. For instance, one could exploit the Cayley transform to determine the search curve, such as in the algorithms proposed in \cite{wen2013,zhu2017}. However, as explained in \cite{tonin2020}, these methods could be numerically problematic because determining the search curve requires a matrix inversion at each iteration. Therefore, we propose to employ the Projected Gradient Descent (PGD) algorithm, which specifies an iterative algorithm projecting $H$ onto the Stiefel manifold at each iteration $k$.
The iterates for minimizing \eqref{eq:nlevels} are specified by
\begin{align}
	H^{k+1} &= \mathbf{\Pi}_{\text{St}(s,N)}(H^k - \alpha_k \nabla J_2 (H^k,w^k,b^k)),\\
	w^{k+1} &= w^k - \alpha_k \nabla J_{\nlevels} (H^k,w^k,b^k),\\
	b^{k+1} &= b^k - \alpha_k \nabla J_{\nlevels} (H^k,w^k,b^k),
\end{align}
where $\mathbf{\Pi}_{\text{St}(s,N)}$ is the Euclidean projection onto the Stiefel manifold and $\alpha_k$ is the stepsize selected via backtracking.
This projection is computed using the compact SVD of $H^k$, while the weights $w$ and bias $b$ of the LSSVM classifier level are not projected onto the Stiefel manifold.  %
The objective \eqref{eq:mlp} with MLP classifier is also optimized with PGD, where the weights of the MLP are not projected on the Stiefel manifold but are trained using Adam \cite{adam}.

The variables can be initialized randomly from the standard normal distribution and \textit{end-to-end} training
or with unsupervised initialization with \textit{fine-tuning}. 
In the unsupervised initialization scheme, the KPCA levels are first trained in the DKPCA unsupervised setting \cite{tonin2023} by considering only the unsupervised terms of the objective \eqref{eq:nlevels}; after convergence, the full model is fine-tuned using labels.

\section{Experimental Evaluation}
In this section, we conduct numerical experiments on standard benchmark datasets to evaluate the DRKM classifier and explore its properties. Specifically, we focus on efficient learning for small and large datasets in low- and high-dimensions.

\subsection{Experimental Setup}

\begin{wraptable}{r}{5.9cm}
\vspace{-1.2cm}
    \caption{Tested datasets.}
    \label{table:dataset}
    \centering
    \begin{tabular}{lcccccc}
        \toprule
		Dataset                & $N$   & $N_\text{test}$  & $d$ & $p$ \\ \midrule
		MNIST                  & 4000  & 10000        & $28\times28$ & 10 &\\
		ARCENE                 & 112              & 60        & 10000 & 2 &\\
        Sonar & 166 & 42 & 60 & 2 \\
        Protein & 1489 & 13406 & 357 & 3 \\
        RCV1 & 16193 & 4049 & 47236 & 2\\
        IMDB & 96735 & 24184 & 1002 & 2\\
		\bottomrule
    \end{tabular}
    \vspace{-.5cm}
\end{wraptable}

We compare our method with a standard LSSVM classifier with RBF kernel, and a multilayer perceptron (MLP), and a Convolutional Neural Network (CNN).
We test on MNIST \cite{mnist}, two UCI datasets, ARCENE \cite{arcene} and Sonar, on the bioinformatics dataset Protein, on the RCV1 dataset \cite{libsvm}, and on the IMDB Drama dataset. 
To assess that the proposed DRKM can effectively learn from small amount of training data, we chose a train/test split with few training samples;
details are given in Table \ref{table:dataset}, where $N$ and $N_\text{test}$ are the number of used training and test instances, respectively, $d$ the input dimension, and $p$ the number of classes.
We train all models for a maximum of 100 iterations. In the DRKM objective, we fix $\eta=\eta_j=1, \, j=1,...,\nlevels$ and $\lambda=0.5$. The MLP has one hidden layer with 10 neurons and ReLU activation. 
For the CNN, we employ 2D convolution for 2-dimensional datasets and 1D convolution for 1D datasets.

\subsection{Experimental Results}

\begin{table*}[t]
\caption{Mean classification accuracy (\%) on the test set of MNIST according to the number of KPCA levels. Each column corresponds to a training set size.
}
\centering
\begin{tabular}{lcccccc}
\toprule
Method  & $N=50$ & $N=100$ & $N=250$ & $N=500$ & $N=750$ & $N=1000$ \\ \midrule
1-level & 57.27 (5.75) & 59.76 (4.51) & 71.24 (5.74) & 75.15 (7.47) & 79.68 (7.01) & 80.59 (7.64) \\
2-level & \textbf{61.05} (1.24) & \textbf{65.97} (1.03) & \textbf{75.74} (0.50) & 81.11 (1.22) & 83.89 (0.23) & \textbf{85.29} (0.45) \\
3-level & 61.02 (1.23) & 65.64 (1.70) & 75.69 (0.49) & \textbf{81.35} (0.43) & \textbf{83.90} (0.20) & \textbf{85.29} (0.69) \\ \bottomrule
\end{tabular}
\label{tab:nlevels}
\end{table*}

\begin{table*}[t]
\caption{Classification accuracy (\%) in the small-data regime. %
}
\centering
\begin{tabular}{lcccc}
\toprule
Method            & MNIST & ARCENE & Sonar & Protein \\ \midrule
MLP               & 87.33 & 74.67  & 77.85 & 57.29  \\
CNN               & \textbf{95.80} &  80.33 & 85.71 & 59.87\\
LSSVM            & 91.75  & 79.67 & 88.09 & 57.30  \\
DRKM (DKPCA initialization, no fine-tuning) & 92.72  & 79.67  & 81.95 & 60.13 \\
DRKM (DKPCA initialization, with fine-tuning) & 92.73 & 82.00  & 84.71 & 61.24 \\ 
DRKM (end-to-end training, random initialization)  & 92.78  & \textbf{82.33}  & \textbf{90.27} & \textbf{61.48} \\ \bottomrule
\end{tabular}
\label{tab:competitors}

\end{table*}

\textbf{Do deeper models perform better?} The classification accuracy on the test set of MNIST attained by a DRKM classifier with one, two, and three KPCA levels and one linear LSSVM classification level for multiple $N$ (number of training points) is shown in Table \ref{tab:nlevels}. All models use the same overall number of components: $s_1=6$ for 1-level, $s_1=3$ and $s_2=3$ for 2-level, $s_1=s_2=s_3=2$ for 3-level. The variance is due to random initialization and different selection of the subset of training points. The mean accuracy of the two-level model is higher than that for the one-level model across all $N$, suggesting that the increased depth led to better generalization. For instance, for $N=500$ the mean performance of the one-level model is 75.15\%, while for the two-level model it is 81.11\%. The three-level model performs similarly to the two-level model. The three-level model keeps only two principal components per level, which may not be enough to get improved results. Furthermore, a deeper architecture may be useful on datasets more difficult than MNIST with, for instance, multiple more realistic objects and a complex background, where greater model complexity could boost classification accuracy. 
In general, the variance of the model with one level is significantly larger than that of the models with two and three levels, which means that different initializations have a greater influence on the model with the lowest depth. In conclusion, classification performance can benefit from architectures with multiple levels. \\

\begin{figure}[htbp]
  \begin{minipage}{0.4\textwidth} %
    \centering
    \begin{tabular}{lccc}
    \toprule
    Method            & Protein (full) & RCV1 & IMDB \\ \midrule
    MLP               & 63.01 & 93.38 & 97.88  \\
    CNN               & 67.33 & \textbf{94.83} & \textbf{98.88} \\
    LSSVM             & 58.33 & 94.27 & 97.78 \\
    DRKM              & \textbf{67.61}  & \textbf{94.83} & \textbf{98.88} \\ \bottomrule
    \end{tabular}
    \caption{Classification accuracy (\%) for larger-scale data.}
    \label{tab:largescale}
  \end{minipage}%
  \hspace{5mm}
  \begin{minipage}{0.5\textwidth} %
    \centering
\includegraphics{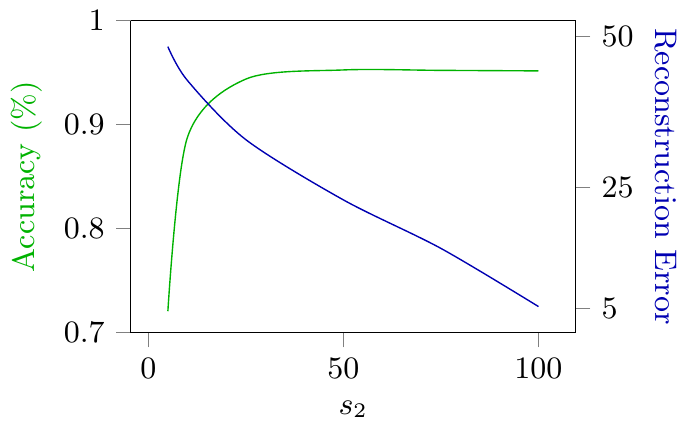}
    \caption{\fra{Test classification accuracy (\%) and training reconstruction error of DRKM on MNIST with varying $s_2$.}}
    \label{fig:s2}
  \end{minipage}
\end{figure}

\noindent \textbf{How does the proposed method perform compared to related techniques?} Table \ref{tab:competitors} compares the classification accuracy (\%) of a two-level DRKM with $s_1=s_2=10$ with MLP classification level against an MLP classifier, a CNN, and a standard nonlinear LSSVM classifier. Results are averaged over five runs. The two-level DRKM outperforms the LSSVM classifier on all datasets, showing that increased depth and end-to-end training can result in better performance. 
In general, the experiments suggest that
the proposed model is well suited for both low-dimensional datasets such as Sonar, as the KPCA levels can derive a higher number of features than the dimension of the input dataset, and high-dimensional datasets such as ARCENE, because the formulations in \eqref{eq:nlevels} and \eqref{eq:mlp} do not depend on the dimension $d$ of the input space thanks to the combination of the dual and primal representations. 
This is not the case for the CNN, which achieves significantly better performance only with spatial data such as the MNIST 2D images.
Table \ref{tab:largescale} evaluates DRKM with end-to-end training on larger datasets, where \textit{Protein (full)} indicates the Protein dataset with $N=11916$ training points. The results show that the proposed method can be applied effectively to both small-scale datasets and larger datasets, for both lower-dimensional data (Protein) and higher-dimensional data (RCV1, IMDB).
\fra{To further evaluate the performance on high-dimensional data, we train DRKM on MNIST for varying number of components $s_2$ keeping fixed $s_1=100$ with cosine kernel. The test classification performance and the training reconstruction error (MSE) are shown in Fig. \ref{fig:s2}. It can be seen that an initial sharp increase in accuracy corresponds to a distinctive drop in reconstruction error; further increasing $s_2$ leads to a decrease in training reconstruction error but small improvement in test accuracy, showing that small values of $s_2$ are enough to achieve good performance on MNIST.}
\\

\noindent\textbf{What is the effect of fine-tuning?} We compare end-to-end training to fine-tuning after DKPCA unsupervised initialization in Table \ref{tab:competitors}.
In addition, we consider a linear classifier trained on the deep features extracted by DKPCA \cite{tonin2023}, i.e., a DRKM with DKPCA initialization but no fine-tuning and a linear classification level.
On Sonar and ARCENE, fine-tuning the hidden units initialized using unsupervised DKPCA initialization results in significantly higher performance compared to simply training a linear classifier on the feature extracted by DKPCA. Note that while random initialization give unfeasible initial points, the unsupervised initialization gives a feasible point. On the other hand, on the more complex MNIST dataset, DKPCA initialization already performs quite well compared to the other initialization schemes. 
While on MNIST the proposed optimization algorithm converges to a similar minimum when the initial $h_i^{(1)}, h_i^{(2)}$ are found by DKPCA, on ARCENE end-to-end training with random initialization considerably outperforms the other initialization schemes, showing that the effect of the initialization schemes is dependent on the dataset at hand. \\ %

\begin{table*}[t]
\centering
\caption{Energy consumption in Wh when training with limited data.}
\resizebox{\textwidth}{!}{
\begin{tabular}{lcccc}
\toprule
Method            & MNIST & ARCENE & Sonar & Protein \\ \midrule
MLP               & 0.0075 & 0.0007  & 0.001 & 0.000002  \\
CNN               & 0.1540 &  0.3746 & 0.036 & 0.00016  \\
LSSVM            & 0.1304  & 0.0002 & 0.00001 & 0.000008  \\
DRKM (DKPCA initialization, no fine-tuning) & 0.0610  & 0.0006 & 0.0009 & 0.000007 \\
DRKM (DKPCA initialization, with fine-tuning) & 1.177 & 0.0044   & 0.021 & 0.0012  \\ 
DRKM (end-to-end training, random initialization)  & 0.316  & 0.0051 & 0.016 & 0.00026 \\ \bottomrule
\end{tabular}
}
\label{tab:co2}
\end{table*}

\begin{table*}[t]
\centering
\caption{\fra{Efficiency comparisons on running time and memory consumption on the limited data benchmarks where training time (s) and peak training memory usage (MB) are given.}}
\begin{tabular}{lcccccccc}
\toprule
\multirow{2}{*}{Method\hspace{5mm}} & \multicolumn{4}{c}{Time (s)} & \multicolumn{4}{c}{Memory (MB)} \\ \cmidrule(lr){2-5} \cmidrule(lr){6-9}
& MNIST & ARCENE & Sonar & Protein & MNIST & ARCENE & Sonar & Protein \\ \midrule
MLP               & 1.57 & 0.22  & 0.03 & 0.53 & 12.29 & 22.07 & 2.88 & 8.38 \\
CNN               & 4.55 &  11.36 & 0.33 & 2.88 & 74.46 & 5127.45 & 86.27 & 289.12  \\
LSSVM             & 0.91  & 0.07 & 0.002 & 0.05 & 50.41 & 22.08 & 0.59 & 0.60  \\
DRKM              & 0.98 & 0.15 & 0.07 & 0.42 & 67.89 & 43.52 & 5.24 & 40.53 \\ \bottomrule
\end{tabular}
\label{tab:comp:small}
\end{table*}

\begin{table*}[t]
\centering
\caption{\fra{Efficiency comparisons on running time and memory consumption on the larger-scale benchmarks.}}
\begin{tabular}{lcccccc}
\toprule
\multirow{2}{*}{Method\hspace{5mm}} & \multicolumn{3}{c}{Time (s)} & \multicolumn{3}{c}{Memory (MB)} \\\cmidrule(lr){2-4} \cmidrule(lr){5-7}
& Protein (full) & RCV1 & IMDB & Protein (full) & RCV1 & IMDB\\ \midrule
MLP               & 5.26 & 5.92  & 23.80 & 10.32 & 5.18 & 32.44  \\
CNN               & 25.79 & 99.18 & 372.64 & 326.43 & 720.43 & 700.12 \\
LSSVM             & 0.95  & 1.42 & 8.26 & 10.88 & 16.68 & 19.26 \\
DRKM              & 16.46  & 33.72 & 365.17 & 139.20 & 148.78 & 1343.54  \\ \bottomrule
\end{tabular}
\label{tab:comp:large}
\end{table*}

\noindent\textbf{What is the environmental impact?} The experiments above show that the combination of unsupervised KPCA in dual and supervised level in primal is effective when training with limited data.
In efficient AI, it is also important to consider computational efficiency and the corresponding environmental impact of Machine Learning algorithms.
In Table \ref{tab:co2}, we report the energy consumption in the small-data regime in Wh tracked using CodeCarbon.io\footnote{\url{https://codecarbon.io/}}.
Overall, DRKM has comparable or lower energy consumption than CNN while showing higher performance on the 1D datasets. The evaluations on the larger-scale data are presented in Table \ref{tab:co2:large}, where DRKM shows lower energy consumption than CNN while maintaining competitive performance.

\begin{wraptable}{r}{5.5cm}
\vspace{-.8cm}
    \caption{Energy consumption (Wh) for larger-scale data.}
    \vspace{-.2cm}
    \label{tab:co2:large}
    \centering
    \begin{tabular}{lccc}
    \toprule
    Method            & Protein (full) & RCV1 & IMDB \\ \midrule
    MLP               & 0.020 & 0.029 & 0.076  \\
    CNN               & 0.115 &  0.44 &  2.96 \\
    LSSVM             & 0.0023 & 0.003 & 0.017 \\
    DRKM              & 0.052  & 0.092 & 1.87 \\ \bottomrule
    \end{tabular}
\vspace{-.6cm}
\end{wraptable}
\fra{
We evaluate the efficiency of DRKM by comparing in terms of running time and memory consumption with three commonly used methods: MLP, CNN, and LSSVM. The results are presented in Tables \ref{tab:comp:small} and \ref{tab:comp:large}. 
Table \ref{tab:comp:small} shows the efficiency comparisons when limited data is available, where we measured the training time (in seconds) and peak training memory usage (in megabytes). We can observe that the combination of non-parametric and parametric levels makes it possible for the DRKM to achieve competitive results, with significantly lower memory consumption compared to CNN. 
For instance, in the high-dimensional ARCENE datasets, the DRKM uses 43.52 MB compared to 5127.45 MB used by CNN. The significant advantage in memory usage is due to the dual KPCA levels, whose dimension is independent of the input dimensions. The KPCA levels can perform dimensionality reduction such that the parametric classification level in the DRKM operates from an input space of much lower dimension.
This indicates that our DRKM method strikes a balance between computational efficiency and memory utilization, making it a favorable choice compared to the CNN for applications where data has thousands of dimensions but memory resources are limited.
Table \ref{tab:comp:large} presents the efficiency comparisons on larger-scale benchmarks. The results demonstrate that DRKM exhibits favorable efficiency characteristics, with lower training times compared to CNN. Moreover, despite the larger dataset sizes, DRKM maintains a reasonable memory consumption, outperforming CNN in this regard in all tested datasets except the largest one.
Overall, the experimental results demonstrate that the DRKM approach shows competitive efficiency in terms of both energy consumption, running time and memory usage when compared to established methods, suggesting its potential as an effective and resource-efficient solution in Frugal AI applications.
}

\section{Conclusion}

We propose a DRKM classifier based on the DKPCA representation learning method in its dual representation combined with a primal classification level expressed as LSSVM or as MLP. 
\fra{Our proposed DRKM classifier exhibits several advantageous characteristics that make it well-suited for Frugal AI. 
Firstly, the integration of unsupervised KPCA levels in our model enables superior performance compared to both LSSVM and MLP/CNN when dealing with tabular classification datasets with limited training points. This indicates the ability of our model to extract meaningful features from scarce data, making it particularly useful in resource-constrained scenarios. 
Moreover, our study demonstrates that employing multiple KPCA levels in the DRKM classifier surpasses the performance of using a single KPCA level. This finding highlights the significance of incorporating multiple perspectives in representation learning, leading to enhanced classification accuracy.
Finally, our method showcases versatility in handling datasets of varying input dimensions and number of samples. We show that DRKM has lower energy consumption than the CNN on a number of higher-dimensional UCI datasets thanks to the employed dual representation, whose dimension is independent of the number of inputs. The combination of dual and primal components of the DRKM allows for efficient learning across different data settings, as the kernel-based levels are better suited for high-dimensional data and the parametric classification level is suitable for larger number of data.
Overall, the above findings demonstrate the suitability of the DRKM classifier for Frugal AI applications, where limited resources and diverse data dimensions are common challenges. By leveraging its dual and primal components, our model demonstrates promising capabilities in achieving good performance and efficient training in Frugal AI scenarios.}
\fra{Future work can investigate the performance of the DRKM classifier with fixed-size methods \cite{suykens2002} to further improve its scalability on very large-scale datasets with millions of samples, such as ImageNet. For computer vision tasks, it would also be interesting to explore the application of specific kernels to enhance its performance on image data.}

\scriptsize{
\section*{Acknowledgements} \label{sec:ack}
This work is jointly supported by ERC Advanced Grant E-DUALITY (787960), iBOF project Tensor Tools for Taming the Curse (3E221427), Research Council KU Leuven: Optimization
framework for deep kernel machines C14/18/068, KU Leuven Grant CoE PFV/10/002, and Grant  FWO G0A4917N, EU H2020 ICT-48 Network TAILOR (Foundations of Trustworthy AI - Integrating Reasoning, Learning and Optimization), and the Flemish Government (AI Research Program), and Leuven.AI Institute. This work was also supported by the Research Foundation Flanders (FWO) research projects G086518N, G086318N, and G0A0920N; Fonds de la Recherche Scientifique — FNRS and the Fonds Wetenschappelijk Onderzoek — Vlaanderen under EOS Project No. 30468160 (SeLMA). 
}

\bibliographystyle{splncs04}
\bibliography{references}

\end{document}